\pdfoutput=1

\documentclass[11pt]{article}

\usepackage{acl}
\usepackage{color, colortbl}
\definecolor{Gray}{gray}{0.9}

\usepackage{amsmath}
\usepackage{multirow}
\usepackage{graphicx}
\usepackage{pifont}
\usepackage{amssymb}
\usepackage{caption, subcaption}
\usepackage{array}
\usepackage{booktabs}
\usepackage{times}
\usepackage{latexsym}
\usepackage{comment}
\usepackage[T1]{fontenc}

\usepackage[utf8]{inputenc}

\usepackage{microtype}

\title{Knowledge-Aware Conversation Derailment Forecasting Using Graph Convolutional Networks}

\author{Enas Altarawneh \\   York University \\
\texttt{enas@eecs.yorku.ca}  \\ \And 
  Ameeta Agrawal \\
  Portland State University\\
\texttt{ameeta@pdx.edu} \\ \vspace{-15pt}
  \\\AND
  Michael Jenkin \\
  York University\\
\texttt{jenkin@eecs.yorku.ca} \\
  \\\And
  Manos Papagelis \\ 
  York University\\ \texttt{papaggel@eecs.yorku.ca} \\
  }

\begin{document}
\maketitle
\begin{abstract}
{Online conversations are particularly susceptible to derailment, which can manifest itself in the form of toxic communication patterns including disrespectful comments and abuse. Forecasting conversation derailment  predicts signs of derailment in advance enabling proactive moderation of conversations. State-of-the-art approaches to  conversation derailment forecasting sequentially encode conversations and use graph neural networks to model dialogue user dynamics. However, existing graph models are not able to capture complex conversational characteristics such as context propagation and emotional shifts. The use of common sense knowledge enables a model to capture such characteristics, thus improving performance. Following this approach, here we derive commonsense statements  from a knowledge base of dialogue contextual information to enrich a graph neural network classification architecture.  We fuse the multi-source information on utterance into capsules, which are used by a transformer-based forecaster to predict conversation derailment. Our model  captures conversation dynamics and context propagation, outperforming the state-of-the-art models on the CGA and CMV benchmark datasets}

\end{abstract}

\section{Introduction}

The widespread availability of chat or messaging platforms, social media, forums and other online communities has led to an increase in the number of online conversations between individuals and groups. 
In contrast to offline or face-to-face communication,  online conversations often utilize moderation to maintain the integrity of the platform and protect users' privacy and safety \cite{virtual}. Moderation can help to prevent harassment, trolling, hate speech, and other forms of abusive behavior \cite{thirty}. It can also help to prevent and address conversation derailment. Moderation typically takes place in terms of a gating process that reviews users’ statements prior to being posted to the online platform. This ability to review statements prior to posting provides a unique opportunity to review such statements before they impact the online conversation. The critical question becomes one of automatically flagging such statements prior to their publication.

\begin{figure}[t]
     \centering
\includegraphics[width=8cm]{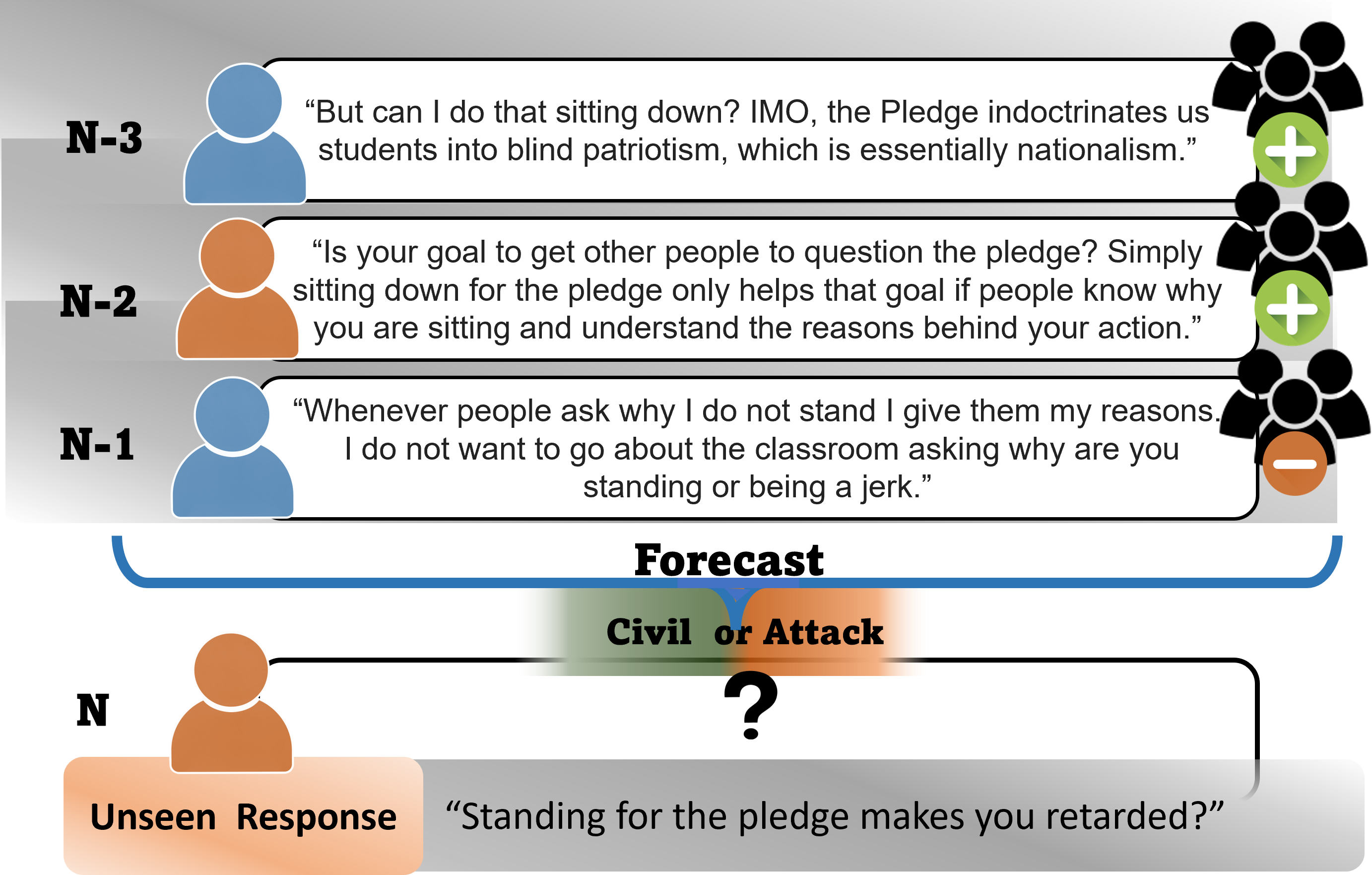}

\caption{A sample conversation from the Reedit change my view (CMV) dataset showing a sequence of text utterances that end with a verbal abuse. Given the conversation context in the previous $N-1$ turns, the task is to predict whether turn $N$ will be a respectful or offensive statement \textbf{prior to it being presented} leading to derailment. The available data for utterances in the CMV dataset contain text, user ID and a public perception negative or positive score  through public votes.}
\label{fig:cmv}
\vspace{-15pt}
\end{figure}

{\em Conversation derailment} refers to the process by which a conversation or discussion is redirected away from its original topic or purpose, typically as a result of inappropriate or off-topic comments or actions by one or more participants. In online conversations, derailment is exacerbated by the lack of nonverbal cues and the perceived anonymity associated with online conversations. Conversation derailment can lead to confusion, frustration, and a lack of productivity or progress in the conversation.
Figure \ref{fig:cmv} shows an example conversation taken from the popular CGA benchmark dataset \cite{CGA}. One can observe that there is offensive language being used in turn $N$ that likely will lead the conversation to derail.  Can we flag the conversation prior to this statement?

In this research, we examine the problem of {\em forecasting conversation derailment}. The ability to predict conversation derailment has multifold benefits: (i) it is {\em more timely}, as it allows for proactive moderation of conversations (before they are published to the conversation and cause any harm) due to early warning, (ii) it is {\em more scalable}, as it allows to automatically monitor large active online communities, a task that is otherwise time-consuming, (iii) it is {\em more cost-effective}, as it may provide enough aid to limit the number of human moderators needed, and (iv) it may identify upsetting content early and prevent human moderators from being exposed to the upsetting content. 

 One effective approach to conversation derailment foresting is a graph-based model that captures multi-party multi-turn dialogue dynamics \cite{altarawneh-etal-2023-conversation}. Although this model out-performs other existing baselines, a limitation of this model is its inability to capture more complex conversational characteristics such as context propagation and emotional shifts. Conversation escalation or derailment is  heavily impacted by the  underlying commonsense knowledge shared between interlocutors. Studies on conversational emotional classification \cite{ghosal-etal-2020-cosmic, zhu-etal-2021-topic}  have shown that incorporating common sense knowledge allows for capturing such conversational characteristics.

Based on these observations, we propose a  novel knowledge aware graph-based model that not only captures multi-party multi-turn dialogue dynamics but is also enriched with common sense knowledge. Thus, allowing the graph model to capture user dynamics, context propagation and mood shifts.  Similar to \citet {altarawneh-etal-2023-conversation}, we leverage other available information associated with conversation utterances. We encapsulate utterance information into capsules that contain text, speaker ID and public perception on whether an utterance is  either being positive or negative. Studies \cite{li-etal-2022-emocaps,higru} have shown that encapsulating utterance information can aid in conversation classification tasks.

Another limitation of \citet{altarawneh-etal-2023-conversation} is the simplicity of the fully connected network classifier used to perform the forecasting task. Inspired by the success of Transformers \cite{ zhu-etal-2021-topic} and to address the limitation in the current stat-of the-art model, we use a Transformer to predict derailment. Derailment of a conversation  depends on the historical dialogue context. For this, We leverage an attention mechanism and the utterance capsules enriched with commonsense knowledge. The major contributions of this work  are:
\begin{itemize}
    \item We develop  a novel graph convolutional neural network model, {\em the  Knowledge Aware Forecasting Graph Convolutional Network (KA-FGCN)}, that captures dialogue user dynamics and leverages common sense  knowledge for derailment forecasting.
    \item Through extensive empirical evaluation we show that KA-FGCN  outperforms state-of-the-art models on the GCA and CMV benchmark datasets. 
   \item The source code of KA-FGCN is publicly available to encourage reproducibility.\footnote{ The GitHub repository will be made available prior to publication.}
\end{itemize}

The remainder of the paper is organized as follows. Section~\ref{sec:related} reviews  related work. The technical problem of interest is presented  formally in Section~\ref{sec:problem}. Section~\ref{sec:models} presents the KA-FGCN. Section~\ref{sec:experiments} presents the experimental setup, the results are presented in  Section~\ref{sec:results}. We conclude in  Section~\ref{sec:conclusions} with a summary and a discussion of ongoing work.

\section{Related Work}
 \label{sec:related}


The CRAFT models introduced by \citet{chang-danescu-niculescu-mizil-2019-trouble} were some of the first models to address the problem of forecasting conversation derailment.  The CRAFT models integrate a generative dialog model that learns to represent conversational dynamics in an unsupervised fashion, and a supervised component that fine-tunes this representation to forecast future events. The performance of forecasting derailment using CRAFT models was improved by incorporating three task-specific loss functions proposed  by \citet{time}. After the rise of  language models  \cite{dynamic} explored how  BERT  \cite{bert}, a pretrained transformer language encoder  can be used for forecasting derailment. This model  resulted in an improved classification F1-score and an earlier forecasting of conversation derailment.  Another  derailment forecasting transformer, Hierarchical-Multi, was proposed by \citet {Yuan_Singh_2023}. Unlike the Bert model, Hierarchical-Multi is a hierarchical transformer-based framework that combines utterance-level and conversation-level information to capture fine-grained contextual semantics. This work also  demonstrates the effectiveness of incorporating multisource information for predicting the derailment of a  outperforming previous models.  FGCN \cite{altarawneh-etal-2023-conversation} also combines utterance-level and conversation-level information and uses multi-source information. Unlike these previous models which were based on  sequence models, FGCN  is built on a graph convolutional neural network that is better able to capture the dynamics of multi-party dialogue, including user relationships and public perception of conversation statements. FGCN performed significantly better than previous sequence models. But it lacked mechanisms to  capture conversation characteristics such as context propagation and mood shifts. In this work we propose to leverage the gains of the graph-based model capturing not  only  multi-party multi-turn dialogue dynamics but also  context propagation and mood shifts by enriching the model with common sense knowledge. In addition, we infuse available utterance information in utterance capsules that contain text, speaker ID, common sense knowledge and public perception on whether an utterance is perceived as either being positive or negative. 
Studies \cite{li-etal-2022-emocaps,higru} have shown that encapsulating utterance information can aid in conversation classification tasks. The model proposed here also uses a Transformer classifier instead of  a the simple classifier used in FGCN, allowing the model to capture long distance context encoded in the conversation representations. 

\begin{figure*}[h]

   \centering\includegraphics[width= 16cm]{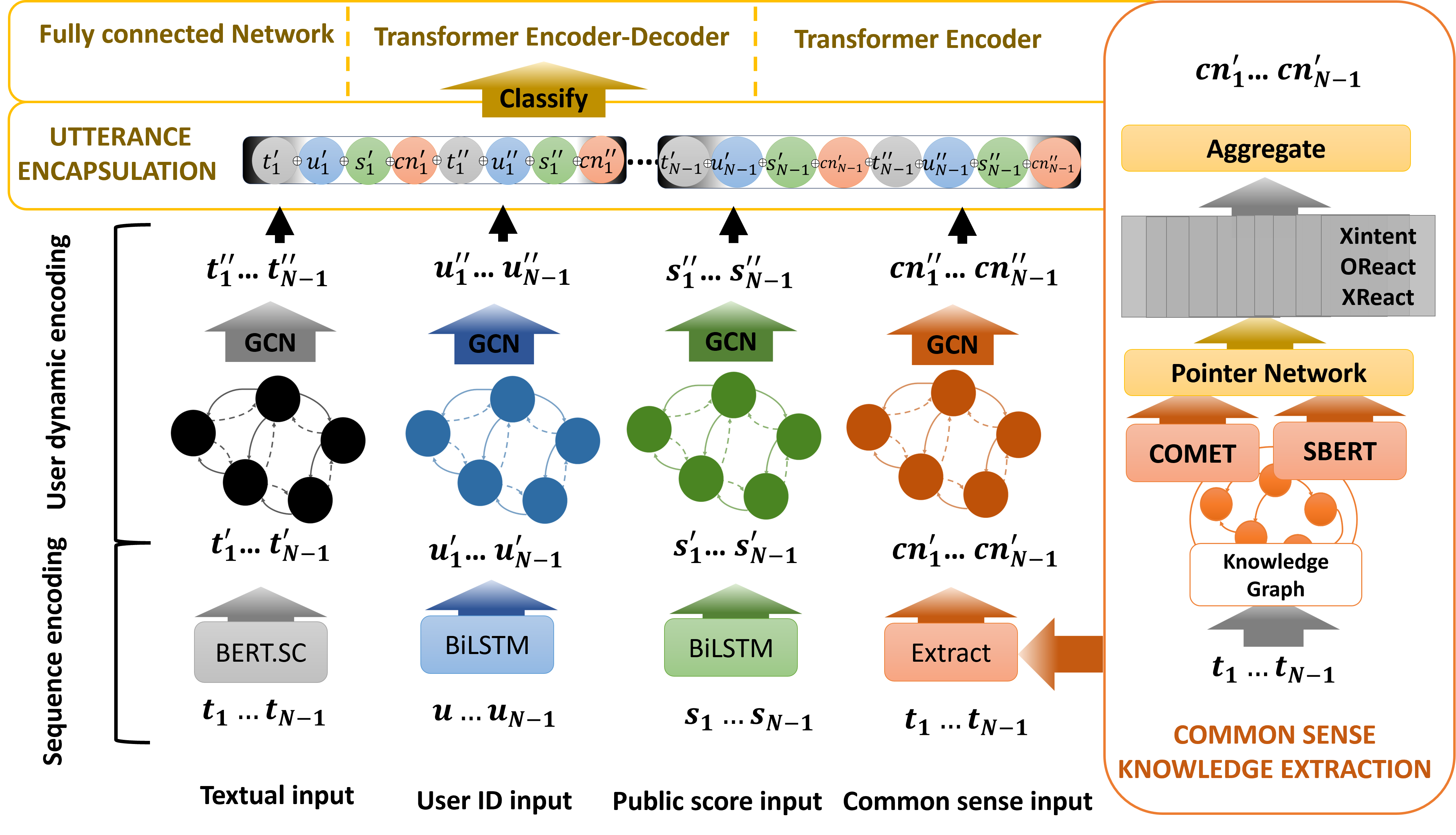}
   \caption{An overview of the Knowledge Aware Graph convolutional Network forecasting model KA-GCN }
   \label{GCN}
  \vspace{-15pt}
\end{figure*}

 There are many commonsense knowledge bases that can  potentially be used to help  inference tasks. For example, ConceptNet \cite{conceptnet}  is a semantic network that contains concept-level relational commonsense knowledge that was used for Emotion Recognition in Conversations.  Event2Mind \cite{rashkin-etal-2018-event2mind} and ATOMIC  \cite{atomic} focus on inferential
knowledge organized as typed if-then relations with variables. COMmonsEnse Transformers (COMET)  \cite{bosselut-etal-2019-comet} is a generative model trained on ConceptNet and ATOMIC, which is able to generate novel, rich, and diverse knowledge that is not in the original knowledge bases. Common sense knowledge has proven to be beneficial for  used in conversation classification tasks \cite{zhong-etal-2019-knowledge, ghosal-etal-2020-cosmic, zhu-etal-2021-topic}. However, Our model will be the first to leverage common sense knowledge for forecasting conversation derailment.

\section{Problem Definition}
 \label{sec:problem}
In this section, we formally define the problem of {\em forecasting conversation derailment}. For a conversation  $\mathcal{C}=\{\{t_1,t_2,...,t_N\}\mathbin{,}\{u_1,u_2,...,u_N\}\mathbin{,}\{s_1,s_2,...,s_N\} \mathbin{,}\{cn_1,cn_2,...,cn_N\}\}$ consisting of $N$ turns, the last turn (i.e., the $N$'th turn) is the potential site of derailment where $l=\{civil,personal$ $ attack\}$ denotes the label of this turn. For the $i$th turn, $t_i$ denotes its text, $u_i$ denotes its user, and $s_i$ denotes an optional score, e.g., number of votes (up-vote/down-vote). An up-vote is a positive impression and a down-vote is a negative impression on the turn utterance. $cn_i$ denotes the extracted utterance common sense knowledge. These types of input are detailed in section \ref{sec:models}. The goal is to forecast the derailment label $l$ of the $N$'th turn given a conversation  $\mathcal{C}$ up to $N-1$ turns (i.e., without any information about the $N$th turn).


\section{The Knowledge Aware Conversation Derailment Forecasting Model}

\label{sec:models}

In this section, we describe KA-FGCN,  visualized in Figure~\ref{GCN}. 

\subsection{Sequential encoding}
The input to the model $X=\{t,u,s, cn\}$ consists of the text $t_i$, user ID $u_i$  the public perception score  $s_i$ and/or common sense knowledge $cn_i$ for each turn in the conversation $i \in \{1, 2, ..., N-1\}$, as described below:

    \noindent \textbf{Textual input} --- the input consists of an encoding of the turn text $t_i$. We vary the text input creating two model variants: KA-FGCN-BRT and KA-FGCN-GPT.  KA-FGCN-BRT uses BERT embedding extracted after fine tuning as described in \citet{dynamic}, resulting in the sequential encoding of the text as the vector $t^{'}_{i}$. KA-FGCN-GPT uses GPT embeddings with a BiLSTM  sequential encoding  resulting in $t^{'}_{i}$.
    
    \noindent \textbf{User ID input} --- the input consists of an encoding of the user ID as a randomly initialized vector $u_i$, where each user has a unique vector; we use  BiLSTM sequential encoding to obtain the utterance user ID vectors $u^{'}_{i}$. We use unique randomly initialized vectors to avoid privacy issues that may arise using actual IDs.
    
  \noindent \textbf{Public perception input} --- the input consists of a popularity score of the number of up-votes on a turn minus the number of down-votes on the same. To obtain the score vector $s_i$ we use equal depth binning to capture three levels of popularity for positive scores and three levels of unpopularity for negative scores. We use  a BiLSTM  sequential encoder on $s_i$  resulting in utterance public perception vectors $s^{'}_{i}$.

   \noindent \textbf{Common sense input } ---  We use ATOMIC \cite{atomic} as a source of external knowledge. In ATOMIC, each node is a phrase describing an event. Edges are relation types linking from one event to another. ATOMIC thus encodes triples such as <event, relation type, event>. There are a total of nine relation types in ATOMIC, of which three are used here: xIntent, the intention of the subject (e.g., ‘to get a raise’), xReact, the reaction of the subject (e.g., ‘be tired’), and oReact, the reaction of the object (e.g., ‘be worried’), since they are defined as the mental states of an event.
   
Given an utterance text $t_i$, we compare it with every node in the knowledge graph, and retrieve the most similar one. SBERT \cite{reimers-gurevych-2019-sentence} is used to compute the similarity between an utterance and events. We extract the top-K events, and obtain their intentions and reactions, which are denoted as $\{e^{sI}_ {i,k}, e^{sR}_{i,k}, e^{oR}_{i,k}\}, k = 1, . . . , K$.

Other than SBERT, We also use the knowledge generation model COMET, which is trained on ATOMIC, to retrieve more common sense knowledge. COMET can take $t_i$ as input and generate the knowledge with the desired event relation types specified (e.g., xIntent, xReact or oReact). The generated knowledge can be unseen in ATOMIC since COMET is essentially a finetuned language model. We use COMET to generate the K most likely events, each with respect to the three event relation types. The produced events are denoted as $\{g^{sI}_{i,k}, g^{sR}_{i,k}, g^{oR}_{i,k}\}, k = 1, . . . , K.$

 With the knowledge retrieved from ATOMIC, we build a pointer network \cite{pointernet} to choose the commonsense knowledge either from SBERT or COMET. The pointer network calculates the probability of choosing the candidate knowledge source as:
 
$P (\Pi(t_i, e_i, g_i) = 1) = \sigma([t_i, e_i, g_i]W_\sigma)$, 

\noindent where $\Pi(t_i, e_i, g_i)$ is an indicator function with value $1$ or $0$, and $\sigma(t) = 1/(1+ exp(-t))$. We envelope $\sigma$ with Gumbel Softmax \cite{JangGP17} to generate the one-hot distribution. The integrated commonsense knowledge is expressed as:

$cn_i = \Pi(t_i, e_i, g_i)e_i +(1 - \Pi(t_i, e_i, g_i))g_i$,

\noindent where $cn_i = \{c^{sI}_{n,k}, c^{sR}_{n,k}, c^{oR}_{n,k}\}^K_k=1$.

With the knowledge source selected, we proceed to select  and aggerate the most informative knowledge using the method described in \cite{zhu-etal-2021-topic} to obtain $cn_i$.

\subsection{Graph Construction} 

For a given conversation, the output of the sequential encoder for each one of the input types $t^{'}_{i}$,  $u^{'}_{i}$, $s^{'}_{i}$ and $cn^{'}_{i}$  is used to initialize the vertices in homogeneous graphs shown in Figure \ref{GCN}. The vertices in the graphs represent the turns in the conversation. Each graph $G_{x} = (V, E, R, W)$, for each type of input $x\in\{t,u,s,cn\}$, is constructed with vertices $v_i  \in V$,  $r_{ij} \in E$ is the labeled edges
 between $v_i$ and $v_j$, the edge labels (relations) $\in R$  and $\alpha_{ij}$ is the weight of the labeled edge $r_{ij}$, with $0 \leq \alpha_{ij}\leq 1$, where $ \alpha_{ij} \in  W$, $i\in \{1, 2, ..., N-1\}$ and $j \in$ the set of all neighboring vertices to $v_i$.
 
For each conversation we construct four types of graphs; $G_x$ for  $ x \in X=\{t,u,s, cn\}$, a text-based $G_t$, a user-based  $G_u$ a public perception score-based $G_s$ and  common sense knowledge $G_{cn}$. Each conversation turn in these graphs is represented as a vertex $v_i \in V $ and is initialized with the sequentially encoded feature vector associated with the graph corresponding data type, for all $i \in \{1, 2, ..., N-1 \}$.


\smallskip\noindent\textbf{ Edge construction}. We establish the direct user to user relationship in a conversation through an edge construction of each graph adopted from \cite{altarawneh-etal-2023-conversation}.  This results in efficient graph modeling with less complexity compared to complete graphs. In user-to-user relationship edge construction, each vertex $v_i$ representing a turn in the conversation has directed edges connecting it to its preceding (parent) and succeeding comments/turns (children). Same user comments (turns) are also connected through directed edges. The user-to-user relation $ \in R$ of an edge $r_{ij}$ is set based on the user-to-user dependency between user $u_{i}$ (of turn $v_i$) and user $u_{j}$ (of turn $v_j$) and is used to label the edge. 

As the graph is directed, two vertices can have edges in both directions with different relations, for example, a forward edge labeled  with user-to-user relation $u_1$-->$u_2$ and a backward edge  labeled  with user-to-user relation $u_2$-->$u_1$. We alos use represent the past (backward) and future (forward) temporal dependency between the vertices.
The edge weights are set using a similarity based attention module. The attention function is computed such that, for each vertex, the incoming set of edges has a sum total weight of 1. The weights are calculated as, $\alpha_{ij} = softmax(v_{i} \odot W_{e}[v_{j_{1}}, . . . , v_{j_{m}} ])$, for $j_k$, where $k= 1, 2, . . , m$, for the $m$ vertices connected to $v_i$, ensuring the vertex $v_i$ receives a total weight contribution of 1.

\subsection{Feature Transformation}
The sequentially encoded text features $t^{'}_{i}$, the user features  $u^{'}_{i}$, the public perception features $s^{'}_{i}$ and the common sense knowledge features $cn^{'}_{i}$ of the graph network are transformed from user dynamic independent into user dynamic dependent feature vectors using a two-step graph convolution process employed by \citet{DGCN,altarawneh-etal-2023-conversation}. In the first step, a new feature vector is computed for each vertex in all four graphs for each input $x \in X=\{t,u,s, cn\}$ by aggregating local neighbourhood information:

\[ x^{''}_{i} = \sigma( \sum\limits_{r\in R} \sum\limits_{j \in N^r_i} \frac{\alpha_{ij}}{c_{i,r}} W_r\, x^{'}_{j}+\alpha_{ii}W_{0}\,x^{'}_{i}), \]

\noindent for $i = 1, 2, . . . , N-1,$ where, $\alpha_{ii}$ and $\alpha_{ij}$ are the edge weights, and $N^r_i$ is the neighbouring indices of vertex $i$ under relation $r \in R$. $c_{i,r}$ is a problem specific normalization constant automatically learned in a gradient based learning setup. $\Sigma$ is an activation function such as ReLU, and $W_{r}$ and $W_{0}$ are learnable parameters of the transformation. In the second step, another local neighbourhood based transformation is applied over the outputs of the first step, as:

\[ x^{''}_{i} = \sigma(  \sum\limits_{j \in N^r_i} W\, x^{''}_{j}+\alpha_{ii}W_{0}   \,x^{''}_{i}),\]

\noindent for $i = 1, 2, . . . , N-1$, where, $W$ and $W_{0}$ are transformation parameters, and $\sigma$ is the activation function. This two step transformation accumulates the normalized sum of the local neighbourhood.

\subsection{The Derailment Forecaster}

To form the final turn or utterance representation, the sequential encoded vectors $t^{'}_{i}$, $u^{'}_{i}$, $s^{'}_{i}$ and $cn^{'}_{i}$, and the user dynamic encoded vectors $t^{''}_{i}$, $u^{''}_{i}$, $s^{''}_{i}$ and $cn^{''}_{i}$ are concatenated for each turn $i$ in a conversation to form an utterance information capsule $d_i$.
 
We vary the  information encapsulated in the utterance capsules to understand the impact of each data type on forecasting derailment, these variants are described in Table \ref{variants}. 

The utterance capsules $d_i$ for $i\in \{1,2....,N-1\}$ are concatenated to form a representation of the conversation $C$:
$$C^{'} = [d_1,d_{2}...,d_{N-1} ]$$

Finally, $C^{'}$ is fed to  a classifier. We vary the types of classifiers in the study to show the impact of each type of classifier on derailment forecasting. These classifiers are described below.

\begin{table}[!t]
\centering
\setlength{\tabcolsep}{1pt}
\begin{tabular}{lccccc} 
\toprule
     
 \textbf{Variant} &\textbf{$t$}&\textbf{$cn$}&\textbf{$u$}&\textbf{$s$} &\textbf{Capsule $d_i$}  \\
\midrule
TCN &\checkmark  &\checkmark & \ding{53} &\ding{53} & $[t^{'}_{i}, cn^{'}_{i},t^{''}_{i},cn^{''}_{i}]$\\
TCNU & \checkmark &\checkmark &\checkmark & \ding{53}  & $[t^{'}_{i}, u^{'}_{i},cn^{'}_{i},t^{''}_{i}, u^{''}_{i},cn^{''}_{i}]$\\
TCNS & \checkmark &\checkmark &\ding{53}&\checkmark & $[t^{'}_{i}, s^{'}_{i},cn^{'}_{i},t^{''}_{i}, s^{''}_{i},cn^{''}_{i}]$ \\
TCNSU & \checkmark &\checkmark &\checkmark & \checkmark& $[t^{'}_{i}, u^{'}_{i},s^{'}_{i},cn^{'}_{i},t^{''}_{i}, u^{''}_{i},s^{''}_{i},cn^{''}_{i}]$\\

\bottomrule
\end{tabular}
\caption{ Variants of the utterance information capsules at classification time. The sequential encoded vectors $t^{'}_{i}$, $u^{'}_{i}$, $s^{'}_{i}$ and $cn^{'}_{i}$, and the user dynamic encoded vectors $t^{''}_{i}$, $u^{''}_{i}$, $s^{''}_{i}$ and $cn^{''}_{i}$ are concatenated for each turn $i$ in a conversation to form an utterance information capsule.}\label{variants}
\vspace{-15pt}
\end{table}

\noindent\textbf{Simple  classifier (S)} --$C^{'}$ is fed to  a classifier with a linear layer, a full connected network and a sigmoid activation function, as described by \citet{DGCN}, to obtain the label $\hat{l}$ of each conversation $C$.

\noindent\textbf{Transformer Encoder only (E)} -- this is a transformer encoder classifier. For this classifier we preserve information about the boundaries between turns in $C$ by inserting a [SEP] token between them. One [CLS] token is further added to the start of the input and one [SEP] token to its end. $C^{'}$ is fed to this encoder classifier to  obtain the label $\hat{l}$ of each conversation $C$.

\noindent\textbf{Transformer Encoder-decoder(ED)}-- here we obtain prediction labels $\hat{l}$ for each conversation $C$ at each iteration time $i$. We use a Transformer encoder-decoder to map the conversation utterance sequence to a sequence of derailment prediction labels, optimally this sequence should have the same derailment prediction label at each utterance iteration, that is the conversation true deraliment label. For this classifier each utterance representation is converted to the [CLS] representation. We enforce a masking scheme in the self-attention layer of the encoder to make the classifier predict derailment in an auto-regressive way, entailing that only the past utterances are visible to the encoder. This masking strategy, preventing the query from attending to future keys, suits better a real-world scenario in which the subsequent utterances are unseen when predicting
the derailment at a point of time. As for
the decoder, the output of the previous decoder
block is input as a query to the self-attention layer.  $C^{'}$ is fed to this encoder-decoder classifier. Since we obtain a sequence of predictions at each turn $i$, we use the label predicated the most in the generated label sequence to produce our classification result.

\begin{table}[!t]
\centering
\begin{tabular}{l|ccc|rrr} 
\toprule
\textbf{Dataset} &\textbf{Train} & \textbf{Val} & \textbf{Test} & \multicolumn{3}{c}{\textbf{Input}}  \\
&&&&$t$&$u$&$s$\\
\midrule
CGA& 2508& 840& 840 &\checkmark  &\checkmark & \ding{53} \\
CMV & 4106 & 1368 & 1368& \checkmark &\checkmark &\checkmark \\

\bottomrule
\end{tabular}
\caption{Statistics of the datasets. $t$ denotes text input,  $u$ denotes user ID input and $s$ denotes public perception score input. All splits are balanced between the two classes.}\label{datasets}
\vspace{-15pt}
\end{table}

\begin{table*}[!t]
\centering
\begin{tabular}{p{.3
cm}p{2.8
cm}>{\centering\arraybackslash}p{1cm}>{\centering\arraybackslash}p{1cm}>{\centering\arraybackslash}p{1cm}>{\centering\arraybackslash}p{1cm}>{\centering\arraybackslash}p{.2cm}>{\centering\arraybackslash}p{1cm}>{\centering\arraybackslash}p{1cm}>{\centering\arraybackslash}p{1cm}>{\centering\arraybackslash}p{1cm}} 
\toprule
&&\multicolumn{4}{c}{\textsc{\textbf{CGA}}} &&\multicolumn{4}{c}{\textsc{\textbf{CMV}}}\\ 
\cmidrule(lr){3-6}\cmidrule(lr){8-11}
&\textsc{\textbf{Model}}&Acc& P& R &F1 &&Acc &P &R& F1\\
\bottomrule

\multirow{5}{*}{\rotatebox{90}{ \textsc{\textbf{Static}}}} 
& CRAFT &64.4& 62.7& 71.7 &66.9 &  &60.5& 57.5 &81.3 &67.3\\
&BERT·SC &64.7& 61.5& 79.4 &69.3 & & 62.0& 58.6& 82.8& 68.5\\
&FGCN &66.9 & 63.3&80.2  & 70.8& &64.7& 60.7& 83.3& 70.2\\
&KA-FGCN-GPT &64.8 &61.6 &78.6& 69.1 & &63.5&59.4 &83.1 & 68.2\\
&KA-FGCN-BRT & 67.4&63.7 &81.0& \textbf{71.3}& &66.6 & 62.7& 82.1& \textbf{71.1}\\

\midrule

\multirow{5}{*}{\rotatebox{90}{ \textsc{\textbf{Dynamic}}}}
&BERT·SC+ &64.3& 61.2& 78.9& 68.8& &56.5& 56.0& 73.2& 61.7 \\
&HR-Multi & 65.2 & 62.3 & 76.9 & 68.9& &64.2 & 62.0 & 73.8 &67.4\\
&FGCN+ &65.9 &62.4 & 80.2& 70.2& &63.5&59.7 & 83.1& 69.5\\
&KA-FGCN-GPT+ & 64.1 & 60.9 &  78.6& 68.7& & 59.6& 56.5 & 79.2& 65.9\\
&KA-FGCN-BRT+ &66.7 & 63.0 & 81.0& \textbf{70.9}& &65.0 &61.1 & 82.1& \textbf{69.9}\\

\bottomrule
\end{tabular}
\caption{ Experimental results for forecasting conversation derailment. For FGCN, FCCN+, KA-FGCN and KA-FGCN we report the best results of the  models variants with a simple classifier. Best F1-score are in bold.}\label{f1-score}
\vspace{-10pt}
\end{table*}

\section{Experimental Setup}
 \label{sec:experiments}

\subsection{Datasets}
We use two datasets for the task of forecasting derailment in conversations. Some statistics of the datasets are summarized in Table~\ref{datasets}.

\medskip
\noindent\textbf{The Conversations Gone Awry (CGA}) dataset  \cite{CGA} was extracted from Wikipedia Talk Page conversations. The conversations were sampled from WikiConv \cite{wiki} based on an automatic measure of toxicity that ranges from 0 (not toxic) to 1 (is toxic). A conversation is extracted as a sample of derailment if the $N$th comment in it has a toxicity score higher than 0.6 and all the preceding comments have a score lower than 0.4. Conversations having all comments with a toxicity score below 0.4 are extracted as  samples of non-derailment. This set of conversations is further filtered through manual annotation to determine whether after an initial civil exchange a personal attack occurs from one user towards another. The conversations include the turn with the 
personal attack. This means all $N-1$ turns in a conversation are civil and the $N$'th one is either civil or contains a personal attack. The dataset also contains additional information about each comment in the conversation such as the user posting the comment and the ID  of the parent comment that this comment was a reply to. 

\medskip
\noindent\textbf{The Reddit ChangeMyView (CMV)} dataset \cite{cmv} was extracted from Reddit conversations held under the ChangeMyView subreddit. Conversations  were identified  as derailed if there was a deletion of a turn by the platform moderators. This could have been done under Reddit’s Rule: ``Don’t be rude or hostile to other users.'' Unlike CGA, there is no control to ensure that all the preceding comments to the last one would be civil, resulting in some noise in the data. The dataset also contains additional information about each comment in the conversation such as the user posting the comment, the ID  of the parent comment that this comment was a reply to, and a votes score (i.e., the number of up-votes minus the number of down-votes). 


\subsection{Evaluation Metrics}

Following prior work, we report the performance of the models in terms of accuracy (Acc), precision (P), recall (R), and F1-score. We also report the forecast horizon H introduced by \citet{dynamic}, which is the mean of the turns in which the first detection of derailment occurred for the set of conversations that derail.

\subsection{Implementation}
We use two training paradigms, static and dynamic: In \textbf{static training}, for each conversation we use one training instance with all turns $\{\{t_{1}, ..., t_{N-1}\}$, $\{u_1, ..., u_{N-1}\}$ and/or $\{s_1, ..., s_{N-1}\}\}$ as input. In \textbf{dynamic training}, we use multiple instances of each conversation, by varying the last turn used in each training instance. So, we use $\{t_{1}$, $u_{1}$ and/or $s_{1}\}$  as an instance,  $\{\{t_1, t_2\}$, $\{u_1, u_2\}$ , and/or $\{s_1, s_2\}\}$ as another instance,  and so on until the last instance  $\{\{t_{1}, ..., t_{N-1}\}$, $\{u_1, ..., u_{N-1}\}$ and/or $\{s_1, ..., s_{N-1}\}\}$. So we have $N-1$ instances of each conversation. We denote all dynamically trained models with an added ``+'' at the end of the model name.

At inference time, the model is tested dynamically, i.e., by using turn $\{t_{1}$, $u_{1}$ and/or $s_{1}\}$ as input, and making a prediction $\hat{l_{1}}$, then using turns $\{\{t_1, t_2\}$, $\{u_1, u_2\}$, and/or $\{s_1, s_2\}\}$, and making a prediction $\hat{l_{2}}$, and so on until $N-1$ predictions have been accumulated. The overall predicted label for a given conversation is then obtained as $\hat{l} = \max^{N-1}_{i=1} \hat{l_{i}}$.  

Our models used the same BERT implementation (i.e., \texttt{bert-base-uncased}) as in the baseline models \cite{dynamic}, for our textual sequential encoding, to ensure a comparable evaluation setting. However, it is worth mentioning that any pretrained language model can be used. The graph neural network component described in this work is implemented with settings similar to that reported  by \citet{DGCN}. The results are reported as an average over 10 different runs with random initialization, to account for variance in model performance.  

\subsection{Baselines}
Our KA-FGCN model and its variants are evaluated against the state-of-the-art methods below.

\smallskip\noindent \textbf{CRAFT} \cite{cmv} is a model with a hierarchical recurrent neural network architecture, which integrates a generative dialog model that learns to represent conversational dynamics in an unsupervised fashion, and a supervised component that fine-tunes this representation to forecast future events.
This model is trained statically. 

\smallskip\noindent \textbf{BERT·SC} \cite{dynamic} is a model consisting of the BERT checkpoint with a sequence classification (SC) head, trained statically. 

\smallskip\noindent \textbf{BERT·SC+} \cite{dynamic} similar to BERT·SC consists of the BERT checkpoint with a sequence classification (SC) head, but is instead trained  dynamically. 

\smallskip\noindent \textbf{HR-Multi} \cite{Yuan_Singh_2023}
is a hierarchical transformer-based framework that combines utterance-level and conversation-level information. This model is trained dynamically.

\smallskip\noindent \textbf{FGCN} \cite{altarawneh-etal-2023-conversation} is a graph convolutional neural network and that leverages utterance text, user ID and public perception information. This model is trained statically.

\smallskip\noindent \textbf{FGCN+} \cite{altarawneh-etal-2023-conversation} is a graph convolutional neural network and that leverages utterance text, user ID and public perception information. This model is trained dynamically.

\begin{table}[!t]
\centering
\setlength{\tabcolsep}{3pt}
\begin{tabular}{p{1.1cm}>{\centering\arraybackslash}p{1cm}>{\centering\arraybackslash}p{1cm}>{\centering\arraybackslash}p{1cm}|p{1.1cm}>{\centering\arraybackslash}p{1cm}}  
\toprule
  \multicolumn{4}{c |}{\textsc{\textbf{KA-FGCN}}} &  \multicolumn{2}{c}{\textsc{\textbf{FGCN}}}   \\
  \toprule
 \textbf{Variant} &\textbf{S}&\textbf{E}&\textbf{ED} &\textbf{Variant} & \textbf{S}  \\
\midrule
TCN & 69.3 & 69.8& 69.9 & T &68.3 \\
TCNU & 69.6 & 70.1 & 70.1& TU & 68.8\\
TCNS & 69.7 &70.2 & 70.3  &TS & 69.1\\
TCNSU & 69.9 &70.5 &70.5 & TSU  & 69.5\\

\bottomrule
\end{tabular}
\caption{ Varying the utterance information capsule and classifier for CMV using dynamic training. S, E and ED are the types of classifiers used to obtain the reported F1-score. KA-FGCN  variants have the same data as  FGCN variants with the addition of common sense knowledge.}\label{variants}
\vspace{-15pt}
\end{table}

\section{Results and Discussion} 
 \label{sec:results}

 In the  forecasting conversation derailment experiment we report the results of the static and dynamic training of our model and  compare  with baselines. In the analyzing mean forecast horizon experiment we show how early each model can forecast derailment. 
\subsection{Forecasting Conversation Derailment}

The results in Table \ref{f1-score} show that across both the datasets, our knowledge aware model KA-FGCN, outperforms  baseline models  for both static and dynamic training.  The results show that incorporating common sense knowledge  in the forecasting graph neural network model improves the models' forecasting F1-score. This Indicates the effectiveness of incorporating common sense knowledge for the problem of derailment forecasting. 

The results in  Table \ref{f1-score} confirm  previous studies comparing static and dynamic training  for the problem of deraliment forcaesting\cite{dynamic, altarawneh-etal-2023-conversation}. The results show that  statically trained models outperform their corresponding dynamically trained models. In addition, our dynamically trained  model on the CGA dataest outperform  statically trained baselines. 

\subsection{ Utterance Capsule Information and Classifier Type Sensitivity }
  KA-FGCN results in Table \ref{variants} show that the more  data types included in the utterance capsules the better the model performance. The results also indicate that using a Transformer classifier (E, ED) is better. However,  the encoder-decoder Transformer classifier (ED) only slightly outperforms the encoder only Transformer classifier (E). Indicating that the complexity  of ED  doesn't provide much benefit for this binary classification task. 

  Table \ref{variants} also shows that KA-FGCN outperforms the state-of-the-art  FGCN using the same classifier S. This indicates the importance of incorporating  common sense knowledge to create knowledge aware forecasting models that are able to capture context propagation and mood shifts during the conversation and improve performance.

\begin{table}[t]
\centering
\begin{tabular}{p{2.8cm}>{\centering\arraybackslash}p{1.8cm}>{\centering\arraybackslash}p{1.8cm}} 
\toprule
&\textbf{CGA}&\textbf{CMV}\\ 
\midrule
CRAFT & 2.36& 4.01\\
BERT·SC &2.60& 3.90\\
BERT·SC+ &2.85& 4.06\\
HR-Multi&2.98 &3.78\\
FGCN-T &2.73&  4.03\\
FGCN-T+ &\underline{2.96}&  \underline{4.12}\\
KA-FCCN-BRT &2.90&  4.11\\
KA-FGCN-BRT+ &\textbf{3.02}&  \textbf{4.16}\\
  \bottomrule
\end{tabular}
\caption{ Experimental results of mean forecast horizon (H).The best result is shown in bold. The second best result has been underlined. The + sign denotes dynamically trained models. }\label{Horizon}
\vspace{-10pt}
\end{table}

\subsection{Analyzing Mean Forecast Horizon}

 How early can the model forecast the derailment? To answer this question we calculate the forecast horizon $H$, the mean of the turn in which the first detection of derailment occurred for the set of conversations that derail. A forecast horizon $H$  of 1 means that a derailment coming up on turn $N$ was first detected on turn $N-1$. A longer forecast horizon (i.e., a higher H) allows for earlier interventions and potentially allows moderators to delete the upcoming personal attack as soon as it appears on their platform to avoid any form of escalation. Models that are able to detect a potential intervention earlier have a clear advantage.

In Table \ref{Horizon} we report the results of the mean forecast horizon $H$. The results show that our  knowledge aware graph neural network model KA-FGCN+ with its dynamic training provides the earliest overall forecasting of derailment with a mean H  of 4.16 for CMV, and  3.02 for CGA. Followed by another  dynamically trained model graph model FGCN-T+.   For the statically trained models (CRAFT, BERT.SC, FGCN-T and KA-FGCN), our model KA-FGCN has the best performance as it seems to be able to better model the dynamics between the users of the turns with its graph model by incorporating common sense relations, obtaining a mean H of 4.11 for CMV and,  2.90 for CGA.  





\section {Conclusion}
\label{sec:conclusions}

Unlike previous models that neglect common sense knowledge KA-FGCN is a  knowledge aware graph convolutional neural network forcesting model that is able to capture the dynamics of multi-party dialogue, context propagation and mood shifts.  KA- FGCN performed  better than state-of-the-art models on two widely used benchmark datasets, CGA and CMV. Conversation derailment frequently  impacts our online social interactions. The ability to accurately predict derailment has the potential to enhance the effectiveness of moderation and thus protect individuals who are vulnerable to emotional abuse or harm and improve the overall quality of online interactions.


\section*{Limitations}

Graph models require four or more utterances to form meaningful conversation connections and model their dynamics. In some cases, conversations that derail are not sufficiently long and may be best modeled by simpler sequential models. Any of these models will work best with asynchronous conversations where there is a time lag between the turns to allow for moderation after forecasting.

\section*{Ethics Statement}
In our paper, we focus on the problem of forecast-
ing conversation derailment. The practical employ-
ment of any such system on online platforms has
potential positive impact, but several things would
be important to first consider, including whether
forecasting is fair \cite{ethics1},
how to inform users about the forecasting (in ad-
vance, and when the forecasting affects users), and
finally what other action is taken when derailment
is forecast. Please refer to \cite{ethics2} for a
related overview of such considerations, in the context of abusive language detection.


\bibliography{anthology,custom}
\bibliographystyle{acl_natbib}


\end{document}